\documentclass{article}
\usepackage{spconf,amsmath,graphicx,verbatim}
\usepackage{url,paralist}
\usepackage{placeins}
\usepackage{multirow}
\usepackage{subcaption}
\usepackage{booktabs,threeparttable}
\title{What does a network layer hear? Analyzing hidden representations of End-to-end ASR through speech synthesis}
\name{Chung-Yi Li$^{\star}$ \qquad Pei-Chieh Yuan$^{\star}$ \thanks{$^{\star}$ Two authors contribute equally.} \qquad Hung-Yi Lee}
\address{College of Electrical Engineering and Computer Science, National Taiwan University\\
\small{\texttt{\{r07942080,r07922072,hungyilee\}@ntu.edu.tw}}}

\begin{document}
%\ninept
%
\maketitle
%
% network layer 聽起來怎樣？
\begin{abstract}

%略略改了一下

End-to-end speech recognition systems have achieved competitive results compared to traditional systems. However, the complex transformations involved between layers given highly variable acoustic signals are hard to analyze. In this paper, we present our ASR probing model, which synthesizes speech from hidden representations of end-to-end ASR to examine the information maintain after each layer calculation. 
Listening to the synthesized speech, we observe gradual removal of speaker variability and noise as the layer goes deeper, which aligns with the previous studies on how deep network functions in speech recognition. 
This paper is the first study analyzing the end-to-end speech recognition model by demonstrating what each layer hears. 
Speaker verification and speech enhancement measurements on synthesized speech are also conducted to confirm our observation further.

% we examine the information lost after each layer calculation in end-to-end ASR, by reconstructing input speech from hidden representations.

%Empirical results show that reconstructed speech reveal properties of hidden layers,
%we examine what kind of information is lost after each
%Deep layers 
%We synthesize speech from intermediate representations of ASR. Properties of synthesized speech revealed information present in hidden layers, which is consistent with earlier findings.
%we propose a straightforward approach to examine what kind of information is kept in each layers of end-to-end ASR models. information left in hidden layers
%However, little is understood about the computation involved in the underlying acoustic-to-phonetic transformation given highly variable acoustic signal. 
%that directly transcribe speech to text without any pre-defined alignments
% and examine how information irrelevant to linguistic content, e.g. speaker, noise,  is eliminated in different model architectures.
\end{abstract}
\section{Introduction}
\label{sec:intro} 
%We inspect the amount of speaker information and noise present in hidden states of end-to-end CTC models using simple models trained to reconstruct original speech from hidden states. 
%Opening the black-box of end-to-end ASR models has become an active field of research. Approaches of model interpretation are predominantly done through visualization such as attention plot. We propose a simple yet intuitive method of understanding how hidden layers of end-to-end ASR perceive the input speech.  This makes us Inspecting model internals through sense of hearing, however, has yet to be explored.
%Recently there has been a surge of interest in end-to-end speech recognition systems.

Traditional ASR systems consist of multiple modules: an acoustic model, a language model, and a pronunciation lexicon. Each of them is trained separately and then composed together to perform speech recognition. Complex training procedures are required to obtain state-of-the-art results. End-to-end systems, on the other hand, enjoy several benefits over traditional hybrid systems. The training pipeline is often more straightforward, and each submodule can be optimized jointly to avoid error propagation. State-of-the-art results have also been reported given the vast amount of training data \cite{Park2019}.
 %Lee: 我把 HMM 改成 ASR ，因為 HMM 只有 acoustic model 而已

However, the black-box nature of end-to-end systems prohibits researchers from analyzing the functionality of every single module, e.g., in what layer does the network perform denoising, remove speaker information, or extract linguistic features?
This has led the community to develop various techniques to dissect end-to-end systems, with the hope of better comprehending the complex, highly nonlinear transformations inside the network.
 Previous research analyzing end-to-end ASR involves investigating the underlying phonetic representations learned in the course of training \cite{ DBNAcoustic, nagamine2015exploring, Belinkov2019}. Interpretable filters with SincNet \cite{ravanelli2018interpretable} is proposed and shown capable of removing noise after training. 
%\cite{Wang2017} correlates gate activation signals inside GRU-based autoencoders with phonemes boundaries.
%會不會太少Ｑ

%[重要] 修
Visualization of model internals is a widely considered way to analyze deep networks. 
Examples include visualizing filters in CNNs \cite{zeiler2014visualizing}, plotting saliency maps \cite{Simonyan2013DeepIC}, or drawing alignments learned by attention-based 
 sequence-to-sequence models in end-to-end ASR \cite{chan2016listen}. %Lee: 此處應該要引用論文
Visualization of convolutional kernels is natural for image-related tasks, which led us to wonder whether we can "audify" hidden layers in a similar way, namely, to hear what the layers hear. 
 While the studies above provide fruitful insight regarding the working mechanism of end-to-end systems, approaches utilizing perception other than vision have yet to be discovered.  %把這句挪動位置
 % 會怪怪derㄇ
%[重要] 修

\begin{figure}[t]
    \centering
    \includegraphics[width=0.46\textwidth]{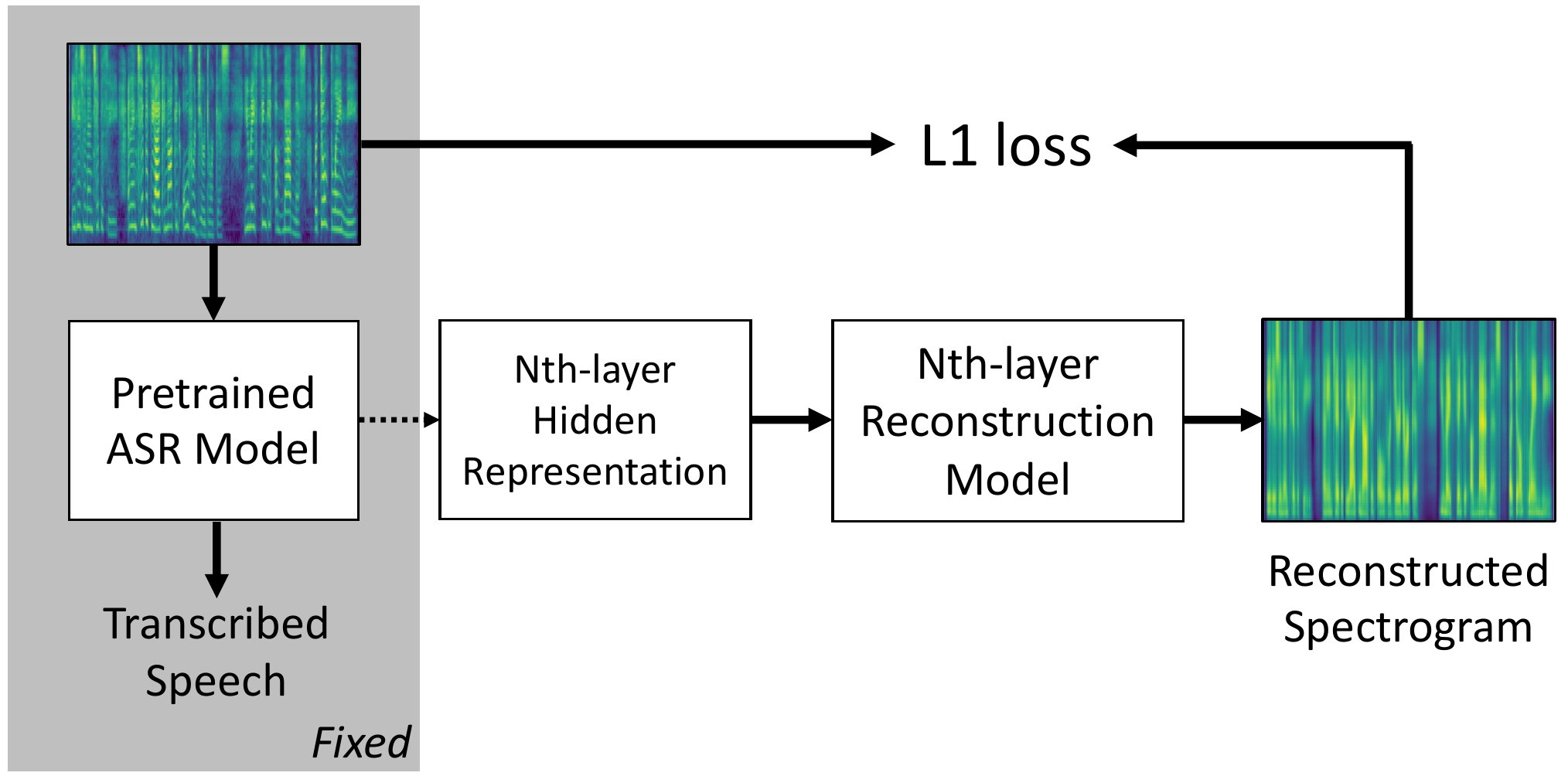}
    \vspace{-3mm}
    \caption{Illustration of the proposed speech reconstruction method applied on hidden representations of the ASR model.}
    \vspace{-2mm}
    \label{fig:model}
\end{figure}

We propose an intuitive and interpretable method of studying what network layers in end-to-end ASR "hear" when transcribing speech.
This is done via reconstructing the input speech from hidden layers of a trained end-to-end ASR system, as shown in figure \ref{fig:model}. 
The experimental results show that the information of speaker characteristics and noise is gradually discarded in each layer, and representations of the same linguistic content spoken by different speakers and under various recording conditions are normalized, which is in line with previous findings \cite{DBNAcoustic}.

To the best of our knowledge, this is the first study to analyze the behavior inside end-to-end ASR models with speech reconstruction from the hidden representations of the network. We strongly recommend the readers to listen to the recovered samples on the demo page \footnote{https://yuanpj.github.io/Voice-in-ASR/}.

%presented visualization of hidden representations learned by Deep Belief Network using t-SNE and show that in deeper layers, the same phonemes spoken by different speakers were clustered together.
%Analyzing hidden layers through synthesizing back into input speech, our study reveals what information is often removed 
%This paper focuses on evaluating representations from different layers of the deep end-to-end speech recognition models  
%This paper focuses on exploring each layer hidden state of 3 kinds of end-to-end ASR models... 
%Our experiment show  

%Our contribution is three-fold:
%\begin{compactitem}
%\item To the best of our knowledge, this is the first study to analyze the behavior inside end-to-end ASR models with speech reconstruction from the hidden representations of the network.
%\item 2
%\item The proposed method reconstructs speech spectrograms from the representations of each different layer of various kinds of end-to-end ASR models.
%\end{compactitem}
%Audio demo page is available at https://yuanpj.github.io/Voice-in-ASR/.
%To the best of our knowledge, this is the first study to analyze the behavior inside the deep neural network architecture, by synthesizing the audio from the hidden representations of end-to-end ASR models.

%methodology

\section{Methodology}
\label{sec:methodology}

%整章要幹嘛，我們先介紹ASR的運作，然後我們提出reconstruct的方式去還原hidden states of ASR，最後我們用一些衡量的方式來衡量ASR model 

\subsection{End-to-end ASR Systems}
To describe our proposed method, we first introduce an end-to-end automatic speech recognition paradigm to be analyzed and define notations in use.

With the goal of investigating different aspect of input signal preserved by the end-to-end ASR in mind, we denote the speech data $x$ to be composed of three different components: the linguistic content $z$, the identity of the speaker $s$ and any other information present in the audio signal $n$ such as recording conditions.
Given the paired training data $x$ and $z$, an end-to-end ASR model can be treated as a function $z = ASR(x)$ which transcribes speech to text.
Further more, for the $k$-th layer of our interest to synthesize speech, we denote the hidden representation $h_{k} = ASR_{k}(x)$.

%End-to-end ASR system usually consists of multiple layers of neural network. Here we denote as $h_{k} = ASR_{k}(x)$ the $k$-th layer output of an end-to-end ASR. We define $x^{t}$ as the input frame of the $t$-th timestep.
%End-to-end systems usually downsample input frames between intermediate layers. Let $d_{k}$ be the down-sampling rate compared to input $x$ at layer $k$, $h_{k}^{t/d_{k}}$ thus refers to the corresponding hidden representations of input frame $x^{t}$ .

%\subsection{Reconstruction Model}
\subsection{Probing Model}
%Lee: 這一節應該要強調，因為 hk 有 loss 資訊，所以無法就算是 minimize reconstruction error ，合成出來的語音也不會一模一樣
% 不然讀者會想說說，用 reconstruction error 去訓練，不是應該跟原來的語音一樣嗎?

% Begin 
In this section, we present our approach of probing model internals through reconstruction.

% how to visualize
With hidden representations $h_{k}$ extracted from the $k$-th layer of ASR, our probing model is trained to recover the input speech using simple feed-forward networks. Formally, let the probing model be $D_{k}$ and $\tilde{x}$ being the synthesized speech, our synthesizer can be trained with the standard reconstruction loss:

\begin{equation}
   \mathcal{L_\text{recon}} = \| \tilde{x} - x \|,\quad
   \tilde{x} = D_{k}(h_{k}).
\end{equation}

Since ASR models extract only the linguistic content $z$ inside the input speech, speaker characteristics $s$ and noises $n$ are discarded in the hidden representations $h_{k}$. The reconstruction loss thus acts as a surrogate for measuring the loss of $s$ and $n$.
Let $\tilde{s}_{k}$ and $\tilde{n}_{k}$ be the amount of information left in $h_{k}$. $\mathcal{L}_{recon} $ is therefore a proxy of the difference between $\{s,n\}$ and $\{\tilde{s}_{k},\tilde{n}_{k}\}$.
% the reason of using your network
 
 Note that our probing model are made non-contextual, and observe simply the current frame representation. We avoid using attention-based models like Tacotron \cite{Tacotron} where every output frame has access to the whole input signal. This ensures that our probing model faithfully convey information left in the intermediate representations learned by the speech-to-text model.

% Expected result of reconstruction 
%An ASR system transforms input audio signal $x$ into transcription $z$. The information of speaker $s$ and recording conditions $b$ is thus unnecessary, and several studies show that speaker variations is often removed at intermediate layers of deep neural networks in an acoustic model \cite{DBNAcoustic}.

\subsection{Measuring the voice of each layer}
 %we describe our approach of measuring the information left in the recovered speech.

 Once we are able to synthesize speech using probing model, we can directly adopt various metrics evaluated on the raw waveform to analyze the intermediate representations of end-to-end ASR. 
 For instance, speaker verification can be conducted to measure the amount of speaker information present in the hidden layers; metrics evaluating performance of speech enhancement models such as perceptual evaluation of speech quality(PESQ)\cite{rix2001perceptual} and short-time objective intelligibility(STOI) \cite{taal2011algorithm} can also be employed to determine the denoising capability of noise-robust ASR.
 % property
 %Denote $s_{k}$ and $n_{k}$ as speaker and noise information left in $h_{k}$, $\tilde{s}_{k}$ and $\tilde{n}_{k}$ as speaker and noise information in $\tilde{x}$. Since $s_{k}$ and $n_{k}$ is unobserved, we treat $\tilde{s}_{k}$ and $\tilde{n}_{k}$ 
%講end to end ASR長怎樣，定義一段speech包含了什麼成分（notation），ASR做到如何抽取重要成分，去蕪存菁的過程
%To measure the amount of speaker information and noise present in the hidden representations of end-to-end speech-to-text models, we first train an end-to-end ASR system on paired audio-text data and freeze their parameters. Then, we feed the speech utterances into the speech-to-text model and collect the hidden states from one layer of the model. These hidden states are then fed to a speech synthesizer which is trained to recover the original speech.
%we train speech synthesizers on the hidden states of speech-to-text model. The goal is to recover the original speech fed into the network from the hidden representations of the model.
% 
%These hyper-parameters diverge from the ASR literature and preprocess our audio signal the same as that of Tacotron \cite{Tacotron}, in order to obtain better synthesized speech for analysis.

\section{Experiments}
\label{sec:experiments}
\subsection{Experimental Setup}
\label{ssec:setup}

\textbf{Data sets}\quad  We use the train-clean-100 subset of LibriSpeech \cite{librispeech} as our clean set for training and the dev-clean and test-clean subset is used for development and evaluation.
We also augment the clean training set with MUSAN\cite{musan} corpus in the settings roughly following that of X-Vector\cite{xvector} as our noisy set and mix the test-clean set with noises at fixed SNR ratio (20, 10, 0dB).

\begin{table}[b]
  \centering
  \resizebox{0.8\columnwidth}{!}{%
  \begin{tabular}{ c  c  c c }
    \toprule
    \multicolumn{1}{c}{\textbf{Model}} & 
    \multicolumn{1}{c}{\textbf{Augmentation}} &
    \multicolumn{2}{c}{\textbf{WER} (\%)}  \\
    \multicolumn{2}{c}{} &
    \multicolumn{1}{c}{dev-clean} &
    \multicolumn{1}{c}{test-clean}\\
    \midrule
    LSTM & No & $30.72$ & $30.98$ \\
    LSTM & Yes & $25.23$ & $25.45$ \\
    VGG-LSTM & No & $29.02$ & $29.55$ \\
    VGG-LSTM & Yes & $22.64$ & $22.51$ \\
    \bottomrule
  \end{tabular}
  }
  \caption{Speech recognition word error rates on LibriSpeech with 100 hours of training data.}
  \label{tab:wer}
\end{table}

\textbf{E2E ASR Models}\quad   We explore two ASR architectures in the experiments. One is a \textbf{VGG-LSTM} model similar to \cite{Hori2017AdvancesIJ}, which is comprised of 4 convolutional layers and 5 bidirectional long short-term memory (LSTM) layers. Each convolutional layer is followed by a ReLU activation layer and max-pooing is applied every 2 convolutional layers.
The other one is a \textbf{pure-LSTM} model, which has only 5 bidirectional LSTM layers, with downsampling performed after 2, 3 and 4-th layer.
The input to the models is 80-dimensional mel-spectrograms.
Deltas and accelerations are added and stacked along the channel dimension.
Both models are trained with Connectionist Temporal Classification(CTC) loss.
Table \ref{tab:wer} shows the achieved WER(\%) of the models.
We take the models trained on clean set without augmentation data as our baseline ASR models, and the ones trained on noisy set are our noise-robust ASR models.

% Reconstruction ?
\textbf{Probing Model}\quad  A 4-layered Highway network is used for probing model of all hidden layers. 
Hidden states of downsampled layers are first up-sampled back by a linear projection then fed into the Highway network. 
The model is trained to minimize L1 loss between the synthesized and the original mel-spectrograms.

\begin{table*}[t]
\centering
\resizebox{1.9\columnwidth}{!}{%
  \begin{tabular}{| c|  l l | c  c  c  c  c  c | c c c c c |}
    \hline
    & & & \multicolumn{6}{c|}{ VGG-LSTM} & \multicolumn{5}{c|}{ pure-LSTM} \\
    %\multicolumn{3}{|c|}{\multirow{2}{*}{BBB}} & \multicolumn{6}{c|}{ VGG-LSTM} & \multicolumn{5}{c|}{ pure-LSTM} \\
    & SNR & type & cnn1 & cnn2 & cnn4 & blstm1 & blstm3 & blstm5 & blstm1 & blstm2 & blstm3 & blstm4 & blstm5\\
    \hline\hline
    (a) & clean & robust &   4.38 &  4.68 & 7.18 & 16.48 &  30.03 &  48.07 &  6.92 &  12.63 & 23.92 &  33.55 &  46.85\\
	(b) & & baseline      &  4.47 &  4.72 &  6.76 &  16.20 & 32.87 & 48.12 & 7.03 & 13.97 &  23.26 & 34.55 & 47.49\\
    \hline
	(c) & 20 dB & robust &  5.80 &  6.27 &  9.23 &  18.11 &  30.40 &  47.75 &    8.44 &  14.97 &  24.98 &  33.80 &  47.07\\
	(d) & & baseline      &  5.80 &   6.19 &   9.21 & 18.85 & 33.28 & 47.82 &   8.84 & 17.65 & 25.37 & 35.53 & 47.54\\
    \hline
	(e) & 10 dB & robust &   9.81 &  10.75 &  13.87 &  20.87 &  30.49 &  47.37 &  13.89 &  19.66 &  25.93 &  33.79 &  46.85\\
	(f) & & baseline      &  9.89 & 10.92 & 15.00 & 24.00 & 36.75 & 47.48 &  14.64 & 24.31 & 30.06 & 37.83 & 47.03\\
    \hline
	(g) & 0 dB & robust   &   23.15 &  24.63 &  28.62 &  31.50 &  34.91 &  46.22 &   29.14 &  33.15 &  33.69 &  38.46 &  46.21\\
	(h) & & baseline      & 23.18 & 25.27 & 29.80 & 37.66 & 44.56 & 48.13 &  30.54 & 39.90 & 42.10 & 45.41 & 47.89\\
%	& Universal & \multicolumn{3}{c|}{76.73}\\
    \hline
  \end{tabular}
  }
  \vspace{-2mm}
\caption{EER results obtained by the ThinResNet\cite{Xie19} on the reconstructed speech utterances.}
\label{tab:eer}
\vspace{-3mm}
\end{table*}
For post-processing, we obtain linear spectrograms by applying pseudo-inverse on generated mel-filterbanks, and then use the Griffin-Lim reconstruction algorithm to generate speech waveform.
%undo normalization done in the pre-processing steps, and then use Griffin-Lim reconstruction algorithm to generate original speech.
%[重要]沒有提到pre-processing要提undo嗎？

%[!重要!]配合methodology想這邊接下來的說詞

%The comparison of the hidden representations is done by measuring the synthesized speech from the layers of VGG-LSTM and pure-LSTM ASR models, both trained on noisy and clean conditions. 

%\input{tables/asr.tex}
%\input{tables/archi.tex}
%Then we get the hidden representations of every layer of the ASR models in both clean and noisy settings, and train the speech synthesizers corresponding to each layer, to recover the original clean or noisy speech.

%\subsection{Visualization of Mel-Spectrograms}
%\label{ssec:visualization}

%這段一定要加，強烈建議去聽 https://yuanpj.github.io/Voice-in-ASR/

%[VoiceId Loss] Since our proposed approach is the first to use only speaker identity for speech enhancement, we strongly recommend the readers to listen to the samples on the demo page1. The inverse short-time Fourier transform is used to generate waveforms with the enhanced magnitudes and the original noisy phase. Note that the objective of the proposed enhancement is for the verification network, not for a human listener. Figure 3 shows example spectrograms.

%\textbf{Visualization of Mel-Spectrograms}\qquad 

%\input{tables/human.tex}
\subsection{Speaker Information}
\label{ssec:spkr}
\textbf{Objective evaluation.}\quad Two speaker verification systems in the previous study are implemented to quantify the amount of speaker information of recovered speech. 
The first model, following \cite{8462665}, which is a LSTM-based neural architecture, maps a sequence of variable-length mel-spectrogram frames to a fixed-dimensional embedding and is trained using the generalized end-to-end(GE2E) loss function.
The second model, ThinResNet\cite{Xie19} is composed of a modified ResNet architecture, which extracts the features from the spectrogram of a speech utterance, and a NetVLAD/GhostVLAD\cite{Zhong18a} layer to aggregate the features along the temporal axis to an embedding.

Both models are trained on the VoxCeleb2\cite{Chung18} dataset, which is comprised of 1 million utterances from 6000 different speakers, and can achieve 4.51\%, 3.34\% EER respectively on the test-clean split of LibriSpeech\cite{librispeech} dataset.

We reconstruct a bunch of speech utterances from every layer of the 4 kinds of ASR models. 
Then we randomly sample the same number of positive and negative pairs to test the above two speaker verification models to estimate the amount of speaker information. 
The results are shown in Table \ref{tab:eer} (We only present the results from ThinResNet because the results from two speaker verification models are quite similar.) The two big columns represent the ASR network architectures, and from left to right is from the top layer to the deepest layer. 
Rows(a, c, e, g) mean that we take the speech at different SNR ratio as the robust VGG-LSTM or pure-LSTM inputs, while rows(b, d, f, h) are to feed the baseline ASR models.
%Lee: Table II 的粗體套的有點奇怪，好像是 noise 和 clean 之間比較好的粗體，可是我們這邊不是正在討論 speaker 嗎???
If we listen to the recovered speech, we can directly recognize that utterances of different speakers from the deeper layer become more indistinguishable from each other than shallower layer. %Lee: 這句的講法奇怪。不是因為  unnatural 所以無法判斷語者，而是因為語者資訊被濾掉了
Meanwhile, the EER values are increased along the layers of all models, meaning that the speaker characteristics are gradually removed out steps by steps.

In comparison between the EER of two network architecture, we found that the cnn part of VGG-LSTM slightly influences the speaker information of hidden representation, while the lstm parts of both architecture seriously wipe out the speaker information.

It's worth noting that although both the baseline and robust ASR models have similar EER when the inputs are clean, robust ASR can achieve better EER under noisy conditions. This could be attributed to the ability of robust models to remove the interference of other speakers in noisy speech.
%Lee: 加一句話說，robust model 比較知道怎麼從 noisy speech 去除 speaker info
%Lee: 下面這兩句是不是有點多餘，反正後面會講到
%We suspect that robust ASR, which is trained using augmented noisy speech, automatically learn the ability to remove noise information.
%We conduct more experiments in the next section to justify our statements.

\label{ssec:human evaluation}
\begin{figure}[b]
    \centering
    \begin{subfigure}[t]{0.23\textwidth}
        \includegraphics[width=\textwidth]{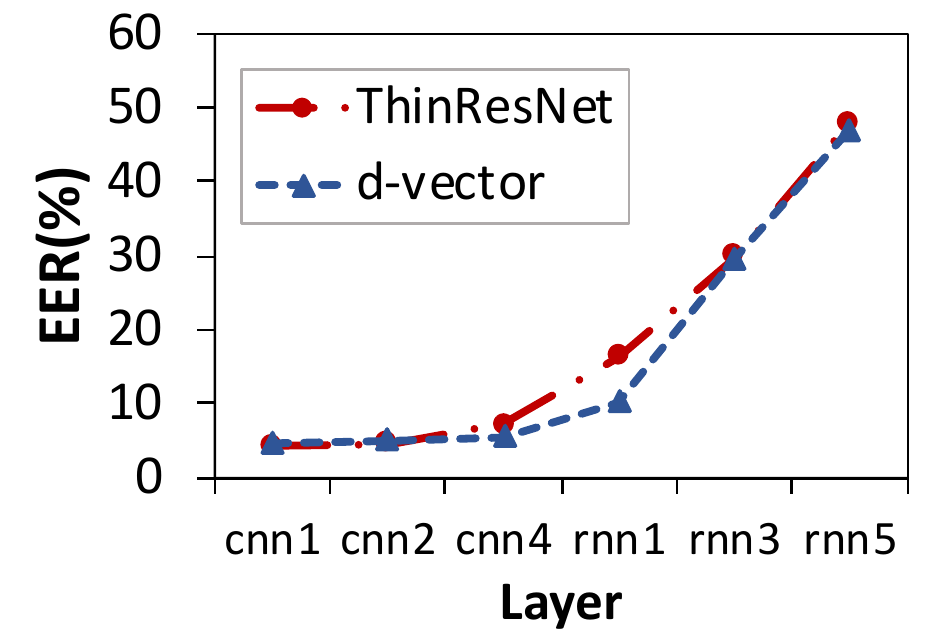}
        \caption{Speaker Verification}
        \label{fig:human_EER}
    \end{subfigure}
    \begin{subfigure}[t]{0.23\textwidth}
        \includegraphics[width=\textwidth]{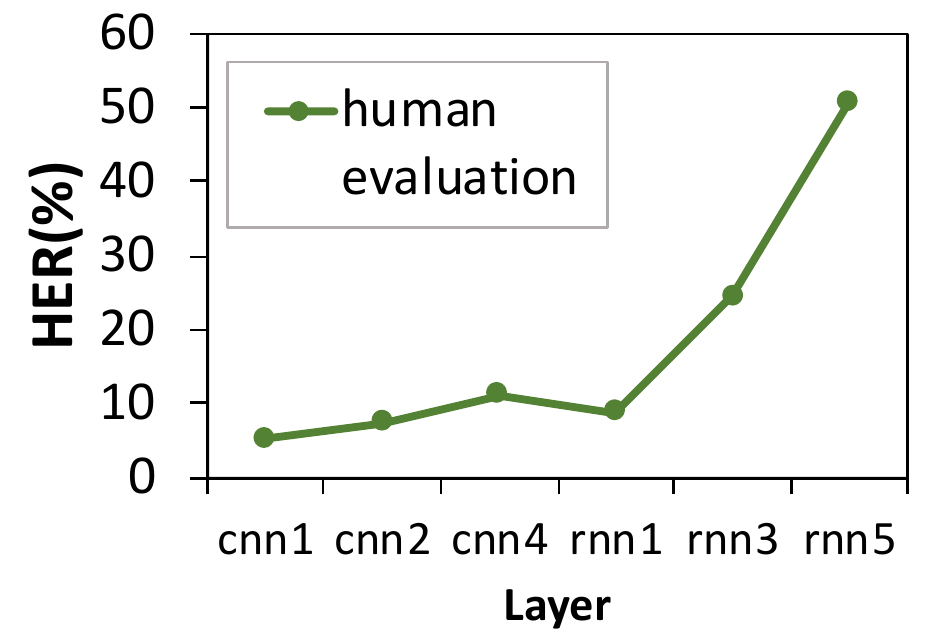}
        \caption{Human Test}
        \label{fig:human_HER}
    \end{subfigure}
    \caption{Evaluation on robust VGG-LSTM ASR}
    \label{fig:human}
\end{figure}
\textbf{Subjective evaluation.}\quad We also perform a human evaluation test to ensure that the measurement of speaker verification models on the reconstructed speech is consistent with that performed by human beings.
%[重要]consistent with ..... <-改名詞
The recovered speech is reconstructed from the hidden representations of well-trained robust VGG-LSTM ASR model.
%====== %Lee: 怎麼突然提到 speaker verification? 怎麼不下一章再講? =========
For the speaker verification measurements, we sample the same number of positive and negative reconstructed speech utterance pairs out of 6 layers, to test the two speaker verification models which will be described in the section \ref{ssec:spkr}. 
%============================================
For the human evaluation test, we randomly sample 10 recovered speech utterance triplets (two from the same speaker, one from another) of 6 layers. Subjects first listen to two speech utterances from different speakers, and then listen to the third one and answer which speaker is same as the third one. 
Human error rate(HER) calculates subjects' mean error rate of the triplets from same layers.
Figure \ref{fig:human} shows the result of the robust VGG-LSTM ASR model. 
In Figure \ref{fig:human_EER}, the EER of both the speaker verification measurement remain low before \texttt{blstm1}, but then the values are dramatically increased at \texttt{blstm3} and \texttt{blstm5}. %Lee: 這是不是應該放在下一節呀?

Results of subjective test Figure \ref{fig:human_HER} are consistent with that of speaker verification models. Most people can verify the correct speaker before \texttt{blstm1}. 
However, the reconstruct utterances from deeper layers are also hard for people to verify, so the HER also greatly rises after \texttt{blstm1}.

%The model is trained to optimize the generalized end-to-end(GE2E) speaker verification loss function, makes the embeddings from the same speaker have high 

\begin{figure*}[ht]
    \centering
    \begin{subfigure}[t]{0.24\textwidth}
        \includegraphics[width=\textwidth]{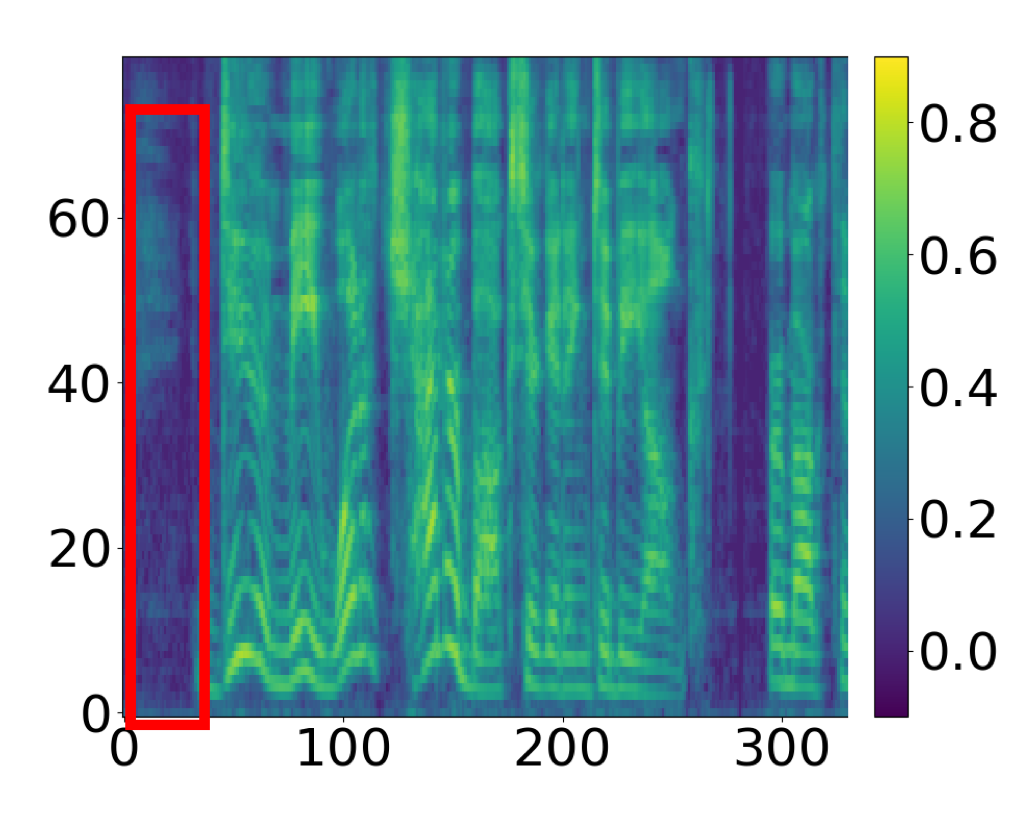}
        \caption{Reference Speech}
        \label{fig:clean_gt}
    \end{subfigure}
    \begin{subfigure}[t]{0.24\textwidth}
        \includegraphics[width=\textwidth]{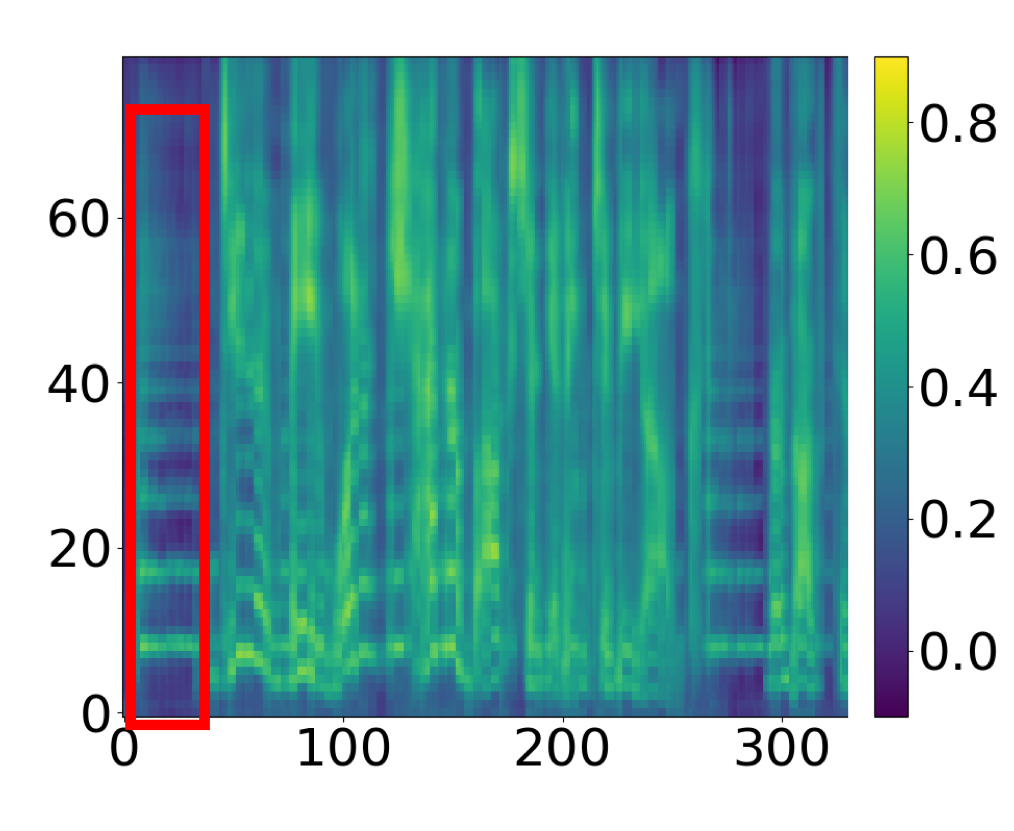}
        \caption{Baseline (\texttt{blstm1})}
        \label{fig:music_nonrobust_1}
    \end{subfigure}
    \begin{subfigure}[t]{0.24\textwidth}
        \includegraphics[width=\textwidth]{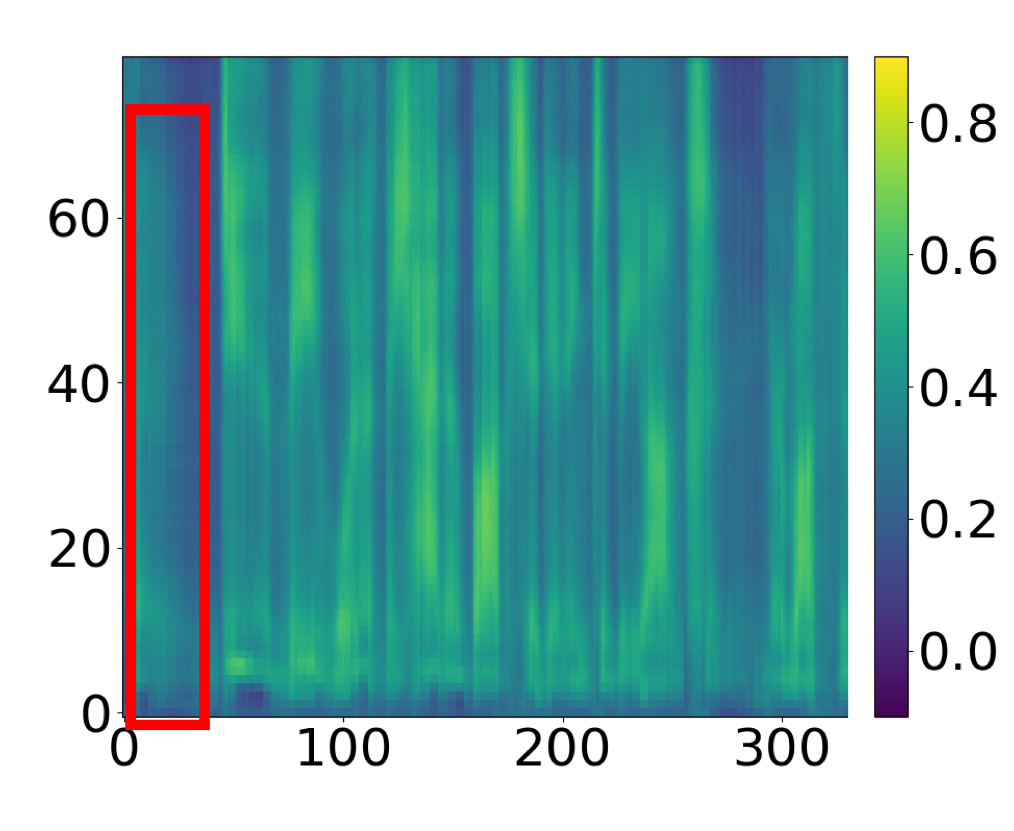}
        \caption{Baseline (\texttt{blstm3})}
        \label{fig:music_nonrobust_2}
    \end{subfigure}
    \begin{subfigure}[t]{0.24\textwidth}
        \includegraphics[width=\textwidth]{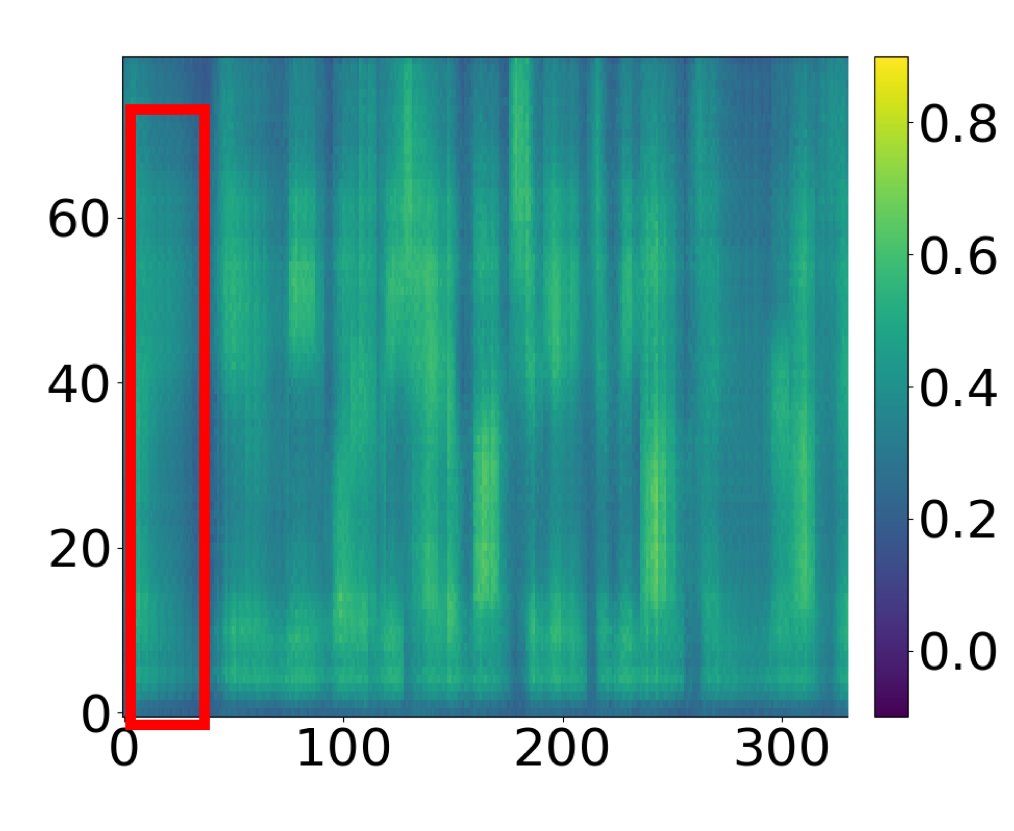}
        \caption{Baseline (\texttt{blstm5})}
        \label{fig:music_nonrobust_3}
    \end{subfigure}
    \vskip\baselineskip
    \begin{subfigure}[t]{0.24\textwidth}
        \includegraphics[width=\textwidth]{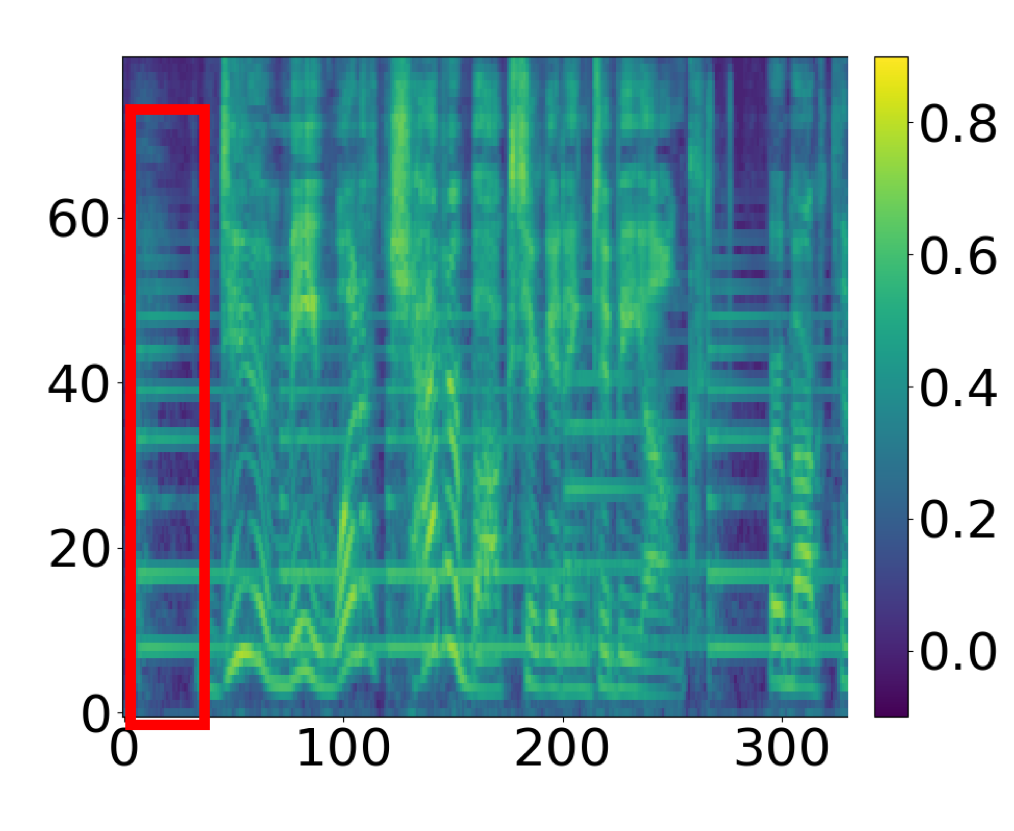}
        \caption{Corrupted Speech}
        \label{fig:music_gt}
    \end{subfigure}
    \begin{subfigure}[t]{0.24\textwidth}
        \includegraphics[width=\textwidth]{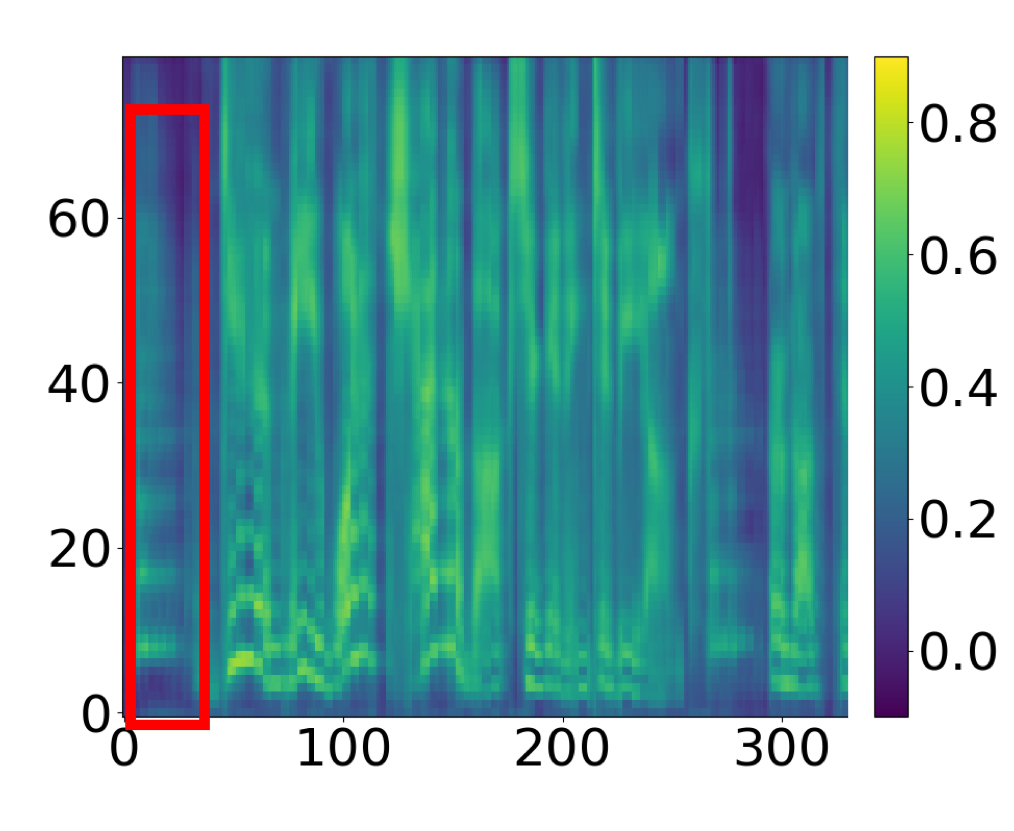}
        \caption{Noise-Robust (\texttt{blstm1})}
        \label{fig:music_robust_4}
    \end{subfigure}
    \begin{subfigure}[t]{0.24\textwidth}
        \includegraphics[width=\textwidth]{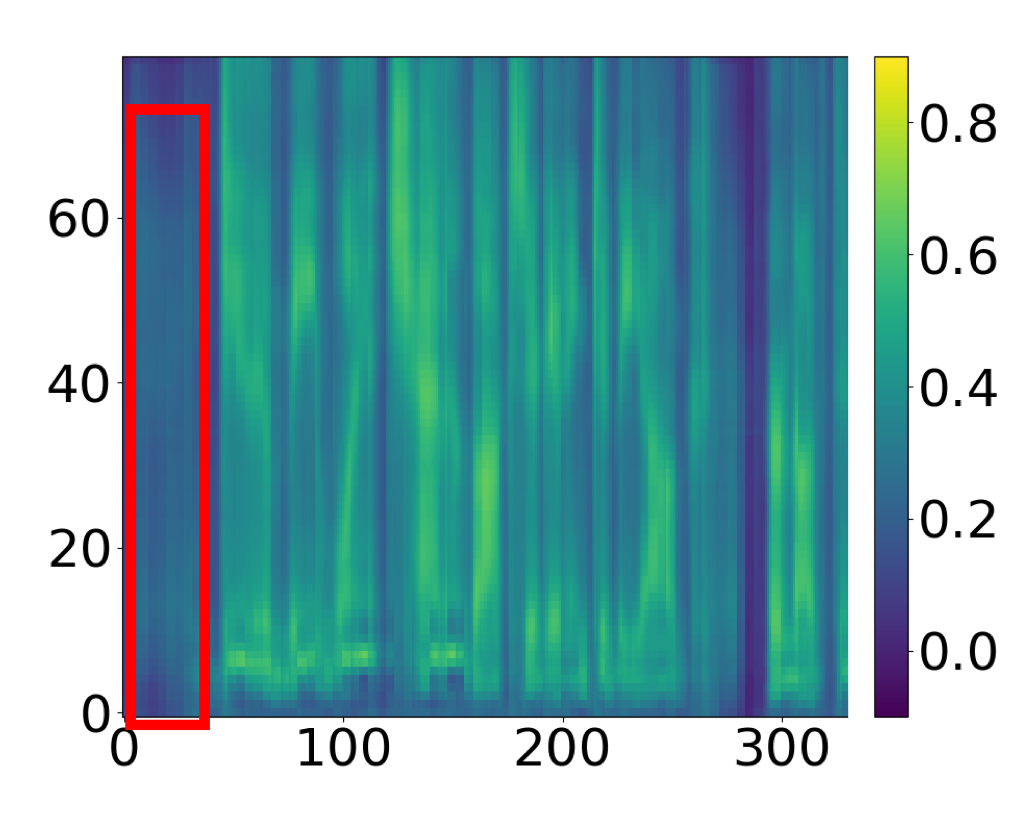}
        \caption{Noise-Robust (\texttt{blstm3})}
        \label{fig:music_robust_5}
    \end{subfigure}
    \begin{subfigure}[t]{0.24\textwidth}
        \includegraphics[width=\textwidth]{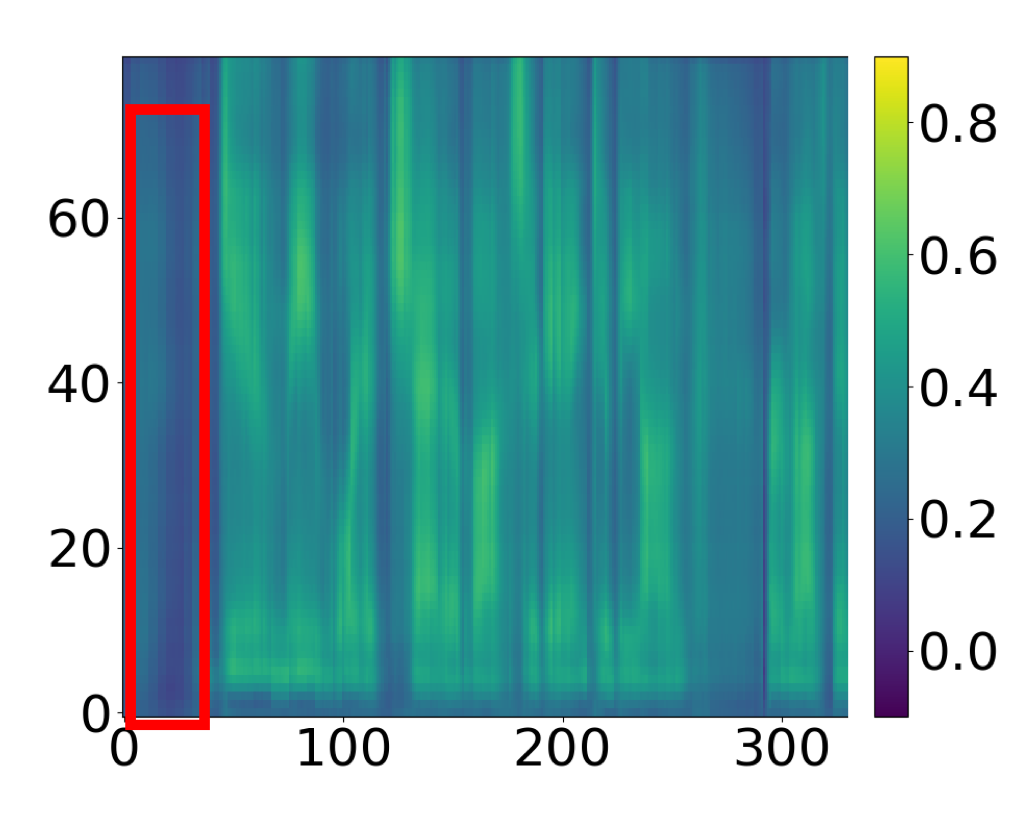}
        \caption{Noise-Robust (\texttt{blstm5})}
        \label{fig:music_robust_6}
    \end{subfigure}
    \caption{80-dimensional mel-spectrograms generated by the probing model of (b)(c)(d) baseline VGG-LSTM and (f)(g)(h) noise-robust VGG-LSTM. Removal of noises (piano music) are most prominent in layer \texttt{blstm1}, highlighted by red bounding boxes. As observed in the figure, noises are still present in (b), but are mostly wiped out in (f).}
    \label{fig:denoise}
\end{figure*}

\begin{figure}[ht]
    \centering
    \includegraphics[width=0.48\textwidth]{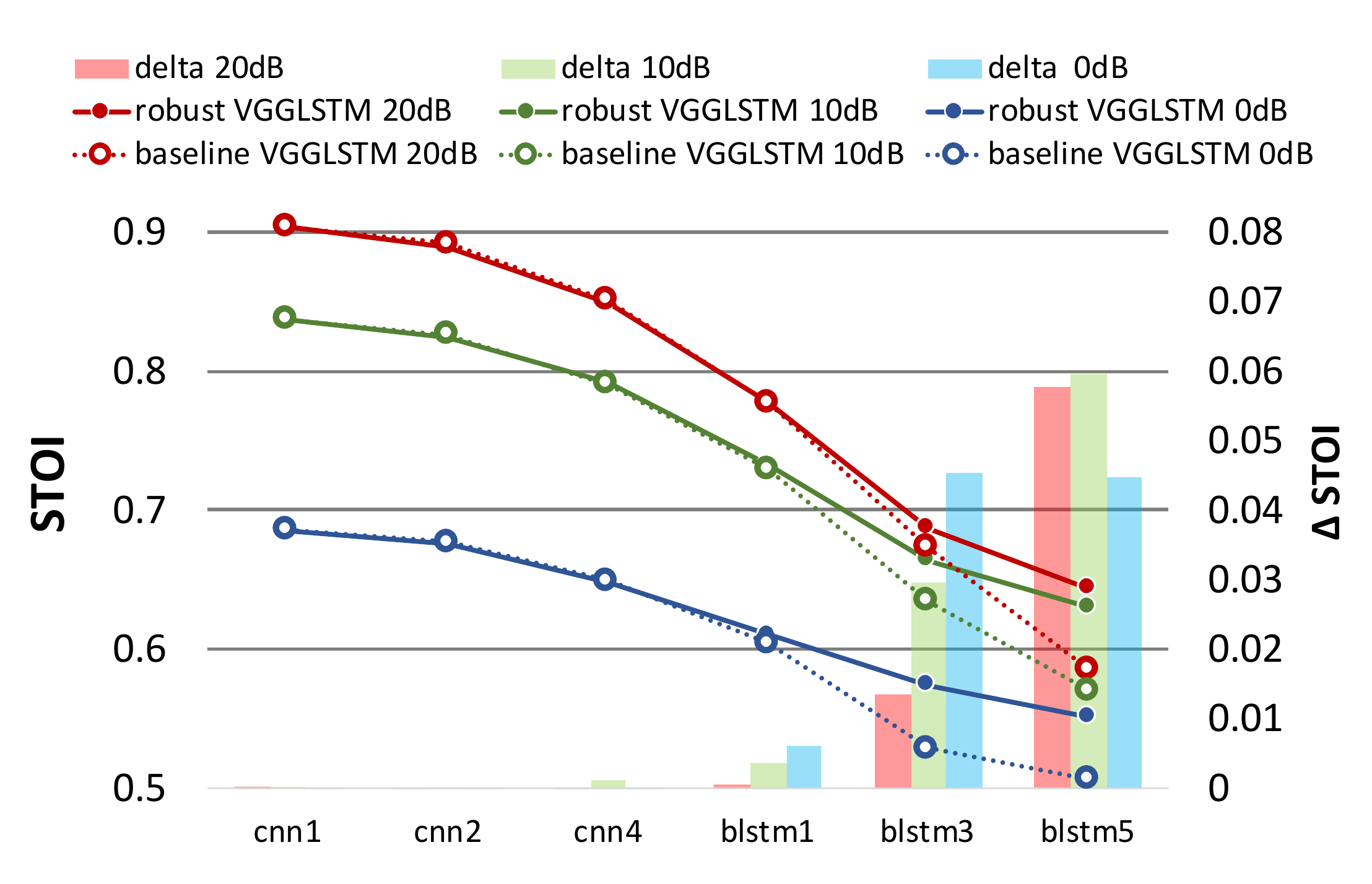}
    \vspace{-3mm}
    \caption{Speech quality measurement of synthesized speech using STOI.}
    \vspace{-2mm}
    \label{fig:stoi}
\end{figure}

% Speech quality measures after enhancement comparing the proposed approach and the DAE
\subsection{Noise Information}
\label{ssec:noise}
In this section, we compare intermediate representations learnt by baseline models and that of their noise-robust counterparts trained on noise-augmented samples. \cite{serdyuk2016invariant,liang2018learning} proposed learning noise-invariant representation for automatic speech recognition, which makes us curious about whether models trained with distorted audio samples automatically learns to draw noisy and clean inputs closer together in the hidden layers.  In Figure \ref{fig:stoi} we report STOI on baseline and robust ASR. We observe an inevitable drop of STOI in both baseline and robust VGG-LSTM. This can be attributed to the loss of speaker information and subsequent degradation of intelligibility. However, the gap of STOI between baseline and robust ASR widens starting from \texttt{blstm1} all the way to \texttt{blstm5}. We suspect that in our models, first few layers of LSTM serves the purpose of removing the background noise information.
Visualization of the mel-spectrograms generated by our probing model further supports the claim. In Figure \ref{fig:denoise}, our probing model, taking the \texttt{blstm1} layer of noise-robust VGG-LSTM as input, which is asked to reconstruct both noisy and clean speech, failed to recover the corrupted speech in figure \ref{fig:music_gt}, demonstrating that noise-robust ASR successfully eliminates added noise in its hidden representation. 

%These qualitative analysis demonstrate that our method is capable of faithfully reconstructing the information left in the hidden states, and 
%To demonstrate that reconstruction method is capable of delivering 

%To quantify the amount of noise present in the hidden layers of end-to-end ASR, We adopt perceptual evaluation of speech quality(PESQ)\cite{rix2001perceptual} and short-time objective intelligibility(STOI) \cite{taal2011algorithm} which is commonly used for evaluating speech enhancement models.

%\input{figs/speaker_spec.tex}

%In this work, we analyzed an E2E speech recognition model in terms of phonetic and graphemic representations. We observed consistent behavior in layer-wise quality across languages, datasets, output labels, and articulatory features. This suggests that such models may benefit from sharing information, for example using multilingual systems as in a recent E2E codeswitching ASR model [35]. In the future, we plan to extend the analysis to other E2E models, such as attentional sequenceto-sequence [19, 20] or purely convolutional models [18].

\section{Conclusion}
\label{sec:conclusion}
In this paper, we seek to answer the question of what a neural network layer hear through directly recovering speech from hidden representations of end-to-end ASR model. We demonstrate that properties of hidden layers can be interpreted and synthesized into the form perceptible by human other than vision, thanks to the task nature of ASR.
Various measures including subjective and objective test are conducted on synthesized speech, and evaluation of speaker variability and noise robustness show findings highly consistent with the literature, demonstrating the effectiveness of the proposed approach.
%\vfill\hfill\pagebreak

\bibliographystyle{IEEEbib}
\bibliography{refs}

\end{document}